# DPAFNet:Dual Path Attention Fusion Network for Single Image Deraining

Bingcai Wei[1], Kunpeng Zhang[1] and Liye Zhang[1*]

[1*]College of Computer Science and Technology, Shandong University of Technology, Zibo City, 255000, Shandong Province, China.

*Corresponding author(s). E-mail(s): zhangliye@sdut.edu.cn;
Contributing authors: weibc1997@gmail.com;
zhangkp0722@gmail.com;

**Abstract**

Rainy weather will have a significant impact on the regular operation of the imaging system. Based on this premise, image rain removal has always been a popular branch of low-level visual tasks, especially methods using deep neural networks. However, most neural networks are but-branched, such as only using convolutional neural networks or Transformers, which is unfavourable for the multi-dimensional fusion of image features. In order to solve this problem, this paper proposes a dual-branch attention fusion network. Firstly, a two-branch network structure is proposed. Secondly, an attention fusion module is proposed to selectively fuse the features extracted by the two branches rather than simply adding them. Finally, complete ablation experiments and sufficient comparison experiments prove the rationality and effectiveness of the proposed method.

# 1 Introduction

Image is an important source of information in modern society, so image quality is particularly important. However, bad weather dramatically affects the quality of images, thus hindering the expected acquisition of information, thereby affecting the normal operation of imaging equipment. Therefore, the restoration of degraded images is essential and has attracted extensive attention from researchers in recent years.

Before data-driven methods show their superiority in the filed of computer vision, model-driven methods are primarily used in single image deraining, including filter-based and prior-based methods in more detail. In recent years, deep learning methods have shone in computer vision and surpassed traditional model-based methods in performance, such as object detection[1], semantic segmentation[2], image classification[3], person reidentification[4]. Among them, convolutional neural networks have been the basic paradigm in the field of computer vision for a long time., until the transformer[5] structure was introduced into computer vision, more and more researchers have devoted themselves to expanding the application of visual transformer[6] in computer vision. Not only in high-level visual tasks, but in low-level visual tasks, such as dynamic scene deblurring of a single image[7], single image dehazing[8], single imagesnow removal[9], and single image denoising[10], data-driven deep learning methods also lead the trend. Therefore, as a branch of image restoration tasks, single image rain removal is also led by deep learning methods. Specifically, Zhang[11] proposed a density-aware multi-stream densely connected network for automatic joint rain density estimation and deraining to determine the rain-density information. Ren[12] proposed a progressive recurrent network, PReNet, as a simple baseline for deraining network. PReNet takes advantage of recurrent computation within stages while reducing the parameters of deep neural networks. Jiang[13] proposed multi-scale progressive fusion network(MSPFN), which can explore the multi-scale collaborative representation for rain streaks by employing recurrent calculation and constructing multi-scale pyramid structure. However, they only use convolutional neural networks for end-to-end mapping, need more attention to particular areas in the image, and thus perform poorly on rain streak removal, as shown in Figure 1. Different from convolutional neural networks, vision transformer[6] can basically not rely on inductive bias, and has stronger long-distance dependence modeling ability, so it can improve the limited attention area of convolutional neural networks to a certain extent. Therefore, in order to solve the problem that the pure convolutional neural network for single image rain removal focuses on limited image regions and lacks global representation ability, we combine vision transformer with convolutional neural network and propose a novel dual-branch attention fusion network for single image rain removal. Specifically, we parallel the vision Transformer module to the residual blocks commonly used in convolutional neural networks and perform feature extraction in different dimensions; we propose a novel attention feature fusion module, which organically fuses the extracted features from the two branches. Ablation studies and comparative experiments verify the rationality and effectiveness of our proposed method.

The main contributions of this work are as follows:

1. We propose a two-branch fusion neural network for single image rain streak removal task. The two branches are the convolutional neural network branch and the vision transformer branch, which can fully exert the local feature modeling ability and long-distance dependence modeling ability of the two branches. At the same time, the features extracted by the two branches are skillfully fused.

2. We propose a novel feature fusion mechanism, which can extract the attention weight from the channel dimension and apply it to the fused CNN branch and Transformer branch, so as to pay more attention to the important feature channels, so as to further effectively fuse the features of the two branches.

3. A large number of ablation experiments prove the rationality of the proposed method. Meanwhile, the performance comparison with various frontier methods on multiple synthetic datasets and multiple real datasets proves the effectiveness of the proposed method.

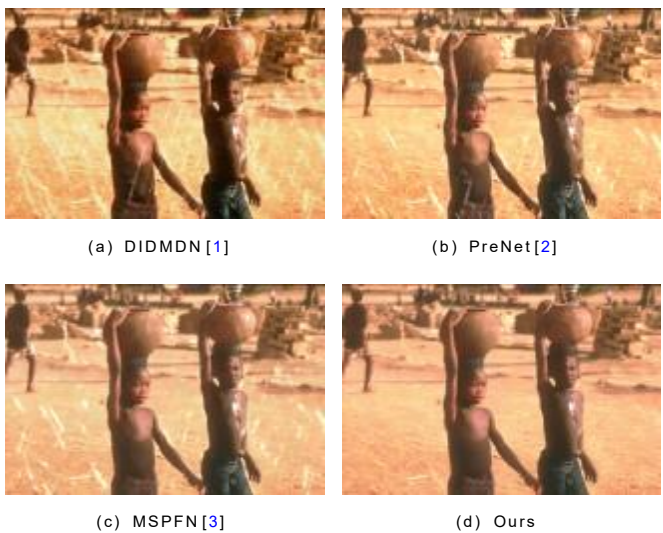

(a) DIDMDN [1]　　　　　(b) PreNet [2]

(c) MSPFN [3]　　　　　(d) Ours

**Fig. 1** As shown in the figure, compared with our method, the effect of removing rain streaks by other methods is insufficient, and their predicted images contain noticeable rain streaks. Our method can remove the rain streak while maintaining the details of the picture

## 2 Releated Works

### 2.1 Rainy Image Formation

Ignoring the complex weather conditions and the intensity of rainfall, animage with raindrops attached can be simply thought of as the sum of rain lines and a clear image. Therefore, in early single-image rain removal studies, the image of rain pattern attachment was simply understood as Formula (1):

$$O = B + \bar{S} \tag{1}$$

where O is the input image with rain streaks, B is the background image, and $\bar{S}$ is the rain streak layer. In order to realistically simulate the rain streak phenomena in real scenes, a new model[14] by accommodating streak accumulation and overlapping rain streaks with different directions was proposed. This model can represent the diversity of rain streaks and can be expressed as Formula (2)

$$O = \alpha\,(B + \sum_{t=1}^{s}\bar{S_t}R) + (1 - \alpha)A \tag{2}$$

where α represents the atmospheric propagation transmittance that is common in image dehazing, $S_t$ is the rain streak layer in the same direction and S is the maximum number of rain streak layers. t is the index of these layers and R represents binary values of 0 or 1, 0 representing areas without rain and 1 representing areas with rain.

### 2.2 Residual Learning and Channel Attention

Training a deeper (such as more than twenty layers) convolution neural network used to be a difficult task[15], because as the number of layers increases, gradients disappear or gradients explode. At the same time, shallow information is transmitted through many layers, and the remaining features are difficult for the network to learn effectively. The emergence of ResNet[16] solves this problem to some extent, introducing residual learning into convolutional neural networks, which greatly deepens the number of layers of trainable networks. By connecting the input directly to the output features after processing by several layers of convolutional neural networks, the network only needs to learn the difference between the output and the input, which reduces the learning difficulty. With the addition of residual blocks, deep convolutional neural networks can also contact shallow features, which makes the propagation of gradients more stable. Many subsequent outstanding neural network architectures use the basic idea of residual blocks, which can be expressed in Formula (3):

$$y = f(x) + x \tag{3}$$

where f(x) represents the convolutional layer in the middle of the residual block, y represents the output of the residual block, and x represents the characteristics of the input.

By mimicking the mechanisms in human vision, researchers have designed many attention mechanisms for computer vision[? ], and the channel attention[18] mechanism is one of the most practical attention mechanisms. By extracting the features of each channel in the previous convolutional layer and further using them in away that applies attention weighting, greater attention can be paid to the more important channels while ignoring some unimportant features. Channel attention can be expressed in Formula (4):

$$F' = M_c(F) \otimes F \qquad (4)$$

where $F'$ are the intermediate feature map and the final refined output, while $\otimes$ denotes element-wise multiplication, in which:

$$M_c(F) = \sigma(W_1(W_0(F^c_{avg})) + W_1(W_0(F^c_{max}))) \qquad (5)$$

where $F^*_{avg}$, $F^*_{max}$ and $W_*$ denote average-pooled features, max-pooled features and CNN's weights, respectively.

## 2.3 Vision Transformer

Transformer[5] has long been a very basic and commonly used model in the field of natural language processing, compared to ordinary recurrent neural networks, it can be calculated in parallel, and at the same time can achieve fairly long-distance modeling. In recent years, more and more researchers have applied the Transformer model to the field of computer vision, and have achieved much better results than convolutional neural networks. The multi-head self-attention(MSA) in the transformer can flatten the loss plane to some extent, and it can be regarded as a low-pass filter compared to convolutional neural networks because it can reduce the variance between feature maps. In, addition, the multi-head self-attention mechanism in the Transformer can be expressed in Equation 6:

$$\text{Attention}(Q,K,V) = \text{softmax}\left(\frac{QK^T}{\sqrt{d_k}}\right)V \qquad (6)$$

where Q, K, V are vectors packed together into three different matrices which are derived from different inputs respectively.

## 3 Proposed Method

In the third chapter, we mainly introduce the content of three parts. Firstly, we introduce in detail the network structure of the two-branch population proposed by us; Then, we introduce our proposed feature fusion mechanism based on channel attention; Finally, we introduce the loss function we use.

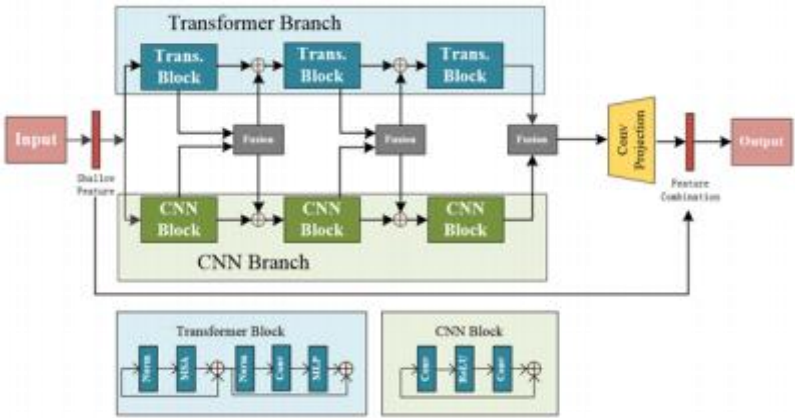

**Fig. 2** The overall structure of DPAFNet proposed in this paper.

## 3.1 Overall Structure

During training, the input image is RGB image. After passing through a convolution layer and a residual block, the obtained shallow features are divided into three directions, one of which is directly fused with the final residual block. One of the other two shallow features enters the Transformer branch and the other enters the CNN branch. After the features of the two branches pass through the corresponding feature refinement module, the features of the channel dimension are organically fused together with the feature fusion mechanism. The structure diagram in Figure 2 describes in detail the encoder part of the entire network structure, that is, the dual-branch feature fusion extraction module. The decoder part of the overall model is called the ' Conv Projection ' part, which consists of four image restoration layers. Each image restoration layer is responsible for sampling the input feature map and performing further feature fitting characterization. Each image restoration layer consists of a bi-linear interpolation upsampling layer, a convolutional layer, a GELU activation function, and a convolutional layer. Finally, the features represented by the whole network fitting are fused with the shallow features extracted by the first layer convolution and residual block.

## 3.2 Channel-Wise Attention Fusion

Convolutional neural networks, such as ResBlock[16], can be regarded as a high-pass filter, because the convolution layer will expand the variance of the input feature map. On the contrary, vision transformer[6], can be regarded as a low-pass filter, which retains low-frequency features while suppressing high-frequency features. Therefore, the mapping frequency of the two to the input features is different, so if the features extracted by the two are simply added, it will affect the performance of the network. We will also mention this in the subsequent ablation experiments. Therefore, we propose a channel-level attention feature fusion mechanism, which includes shallow residual blocks and channel attention blocks. The features of the two branches are first spliced together at the channel level, then mapped by two residual blocks, and finally passed through a channel attention layer. By applying a weight to the features of each channel, the overall network structure can realize the organic fusion of the input features of the CNN branch and the input features of the Transformer branch through the feature fusion module here.

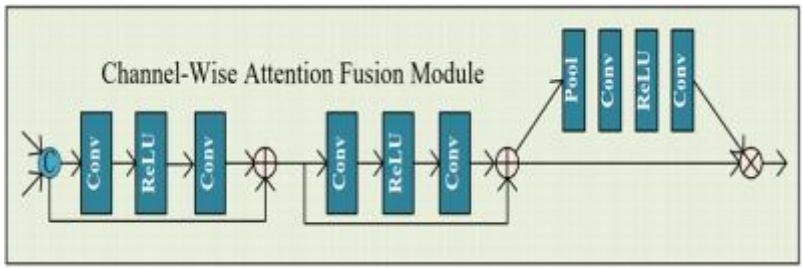

**Fig. 3** The overall structure of Channel-Wise Attention Fusion Module.

### 3.3 Loss Function

The mean square error ( MSE ) loss function is a very commonly used loss function in the field of image restoration. It can reflect the model learning by measuring the Euclidean distance between the predicted value and the label. The MSE loss function can be expressed as Equation 9 :

$$L_{MSE} = \frac{1}{N} \sum_{i=1}^{N} |y - f(x)|^2 \qquad (7)$$

where N represents the number of batches during training,i is the serial number of the current sample, y is the corresponding clear image label, and f(x) is the image predicted by the neural network.

Compared with the MSE loss function, the SSIM(Structural SIMilarity)[19] loss function can guide the training of neural networks in a way that is closer to human visual perception. SSIM can be expressed in equation 10 :

$$L_{SSIM} = 1 - SSIM(f(x), y) \qquad (8)$$

where SSIM represents the structural similarity ( SSIM ) of the two input feature maps.

In order to make the image restored by the neural network have higher authenticity and improve the visual quality of the image, we also use the perceptual loss function[20]. The perceptual loss function measures the performance of the neural network to restore the image by inputting the predicted image and the clear image into the pretrained VGG19 neural network, then taking the feature map of one layer, and then calculating the mean square error loss between the two feature maps. The perceptual loss function[20] is expressed in formula 9.

$$L_{Perp} = \frac{1}{W_{i,j}H_{i,j}} \sum_{x=1}^{W_{i,j}} \sum_{y=1}^{H_{i,j}} \left( \Phi_{i,j}\left(I^S\right)_{x,y} - \Phi_{i,j}\left(G_\Theta{}_G\left(I^B\right)\right)_{x,y} \right)^2 \quad (9)$$

where $\Phi_{i,j}$ is the feature map obtained by the convolution layer within the VGG19[15] and $W_{i,j}H_{i,j}$ are the dimensions of the feature maps.

The rationality of the above loss function is explained in detail in the ablation experiment. Finally, the loss function we use in this paper is shown in Formula 10 :

$$L_{All} = \alpha L_{MSE} + \beta L_{SSIM} + \gamma L_{Perp} \quad (10)$$

# 4 Experimental results and analysis

## 4.1 Experimental details

The ablation experiments and comparative experiments in this paper are carried out on a graphics workstation equipped with NVIDIA RTX3060. The framework used to realize the training and testing of neural networks is Pytorch, and the optimizer used is Adam optimizer. The initial learning rate used in the ablation experiment is $2e^{-4}$. After 200 epoch training, it is attenuated to $1e^{-6}$. At the same time, the batch size is 3. The Rain800[21] dataset is used for training, and the Test100, Rain100H[22] and Rain100L datasets are used for testing. The size of the input image is 128x128.The method proposed in this paper also uses the initial learning rate of $2e^{-4}$ in the comparison experiment. After 200epoch training, it is attenuated to 1e-6. The batch size used is 6, and the size of the input image used is 256. The learning rate warm-up is used to prevent the network from overfitting. During the training process, random horizontal flipping is used for data augmentation to improve the generalization of the network model. This paper uses PSNR and SSIM to quantitatively compare the performance of all the model.

## 4.2 Ablation experiments

Firstly, we conducted ablation experiments on the relevant parts of the network structure proposed in this paper, as shown in Table 1 and Figure 4. It can be seen that the network performance is better when only the visual Transformer branch or the convolutional neural network branch is included, but it is not as good as the network containing two branches. At the same time, the output eigenvalues of the two branches of the network are simply added, and the performance improvement is not as high as the feature fusion mechanism proposed in this paper. Therefore, it is proved that the proposed method is effective in the feature fusion of the two branches.

Then, we performed ablation experiments on the combination of loss functions used in this paper. Starting from the most widely used mean square error loss function, we improve the performance of the proposed neural network on a

Springer Nature 2021 LATEX template

**Table 1** Ablation experiment on network structure.It can be seen that by gradually adding the various methods proposed in this paper, the performance of the network structure in the table is steadily improved, and finally the final structure is achieved.

|  | Test100 PSNR/SSIM | Rain100L PSNR/SSIM | Rain100H PSNR/SSIM |
| --- | --- | --- | --- |
| only Transformer Branch | 22.41/0.775 | 29.49/0.902 | 26.37/0.795 |
| only CNN Branch | 22.10/0.766 | 29.19/0.897 | 26.28/0.797 |
| Transformer+CNN | 22.25/0.776 | 29.95/0.911 | 26.66/0.808 |
| Trans.+CNN+Fusion | **22.37/0.779** | **30.63/0.919** | **26.99/0.814** |

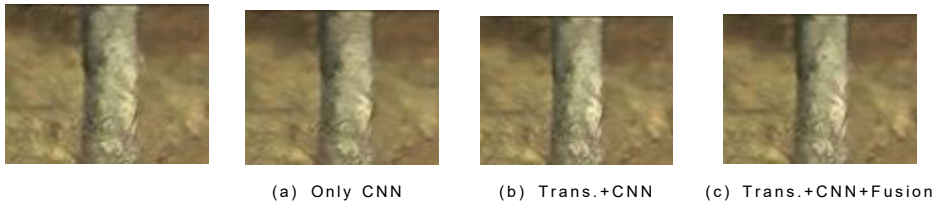

(a) Only CNN    (b) Trans.+CNN    (c) Trans.+CNN+Fusion

**Fig. 4** Comparison of visual effects of ablation experiments on network structure.

**Table 2** Ablation experiment on the choice of loss function.It can be seen that by gradually adding the various methods proposed in this paper, the performance of the network structure in the table is steadily improved, and finally the final structure is achieved.

|  | Test100 PSNR/SSIM | Rain100L PSNR/SSIM | Rain100H PSNR/SSIM |
| --- | --- | --- | --- |
| MSE | 22.37/0.779 | 30.63/0.919 | 26.99/0.814 |
| MSE+Perc. | 22.30/0.788 | 30.65/0.923 | 27.39/0.832 |
| MSE+SSIM | 22.38/0.783 | 30.87/0.921 | 27.20/0.834 |
| MSE+Perc.+SSIM | **22.38/0.790** | **30.92/0.928** | **27.35/0.839** |

single image rain removal task by gradually adding the loss function. As shown in Table 2 and Figure 5, we finally use the combination of mean square error loss function, structural similarity loss function and perceptual loss function as the loss function used in our final method.

### 4.3 Comparative experiments

#### 4.3.1 Synthetic Images

For the comparison experiment with other methods, according to the settings in[23], we use four synthetic data sets for experiments, including Rain14000, Rain1800, Rain800, Rain12, these four training data sets contain a total of 13712 image pairs, we call it MIX. At the same time, this paper compares the proposed method with seven state-of-the-art methods, including DerainNet[24], RESCAN[25], UMRL[26], SEMI[27], DDC[28], DIDMDN[29], PreNet[30] and MSPFN[23]. The experimental results of these methods are provided by [23] , and we use the PSNR and SSIM indicators in scikit-image for reassessment.It can be seen from Figure 6 that compared with the method proposed in this paper, the other methods have the problem of insufficient rain removal or excessive rain removal, the method proposed in this paper can remove the rain streaks to the greatest extent and restore the image closest to the real clear image.

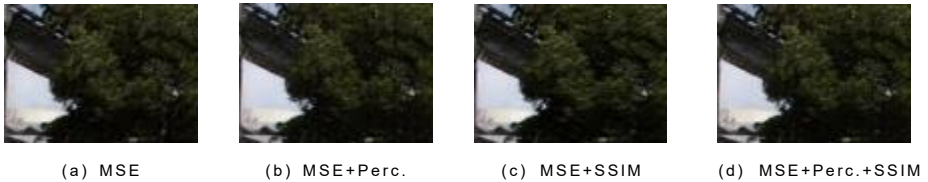

(a) MSE    (b) MSE+Perc.    (c) MSE+SSIM    (d) MSE+Perc.+SSIM

**Fig. 5** Comparison of visual effects of ablation experiments on loss function.

**Table 3** The proposed method is compared with several state-of-the-art methods on multiple test data sets. The results of other methods are derived from Reference 35, and we have remeasured these results, so they may not be the same as the previous indicators.

|  | Test100 PSNR/SSIM | Rain100H PSNR/SSIM | Rain100L PSNR/SSIM | Test1200 PSNR/SSIM |
|---|---|---|---|---|
| DerainNet[24] | 21.90/0.837 | 13.67/0.573 | 26.36/0.873 | 22.24/0.848 |
| DDC[28] | 22.63/0.825 | 27.51/0.884 | 32.77/0.955 | 27.59/0.882 |
| DIDMDN[29] | 21.56/0.811 | 16.31/0.556 | 23.71/0.804 | 27.00/0.883 |
| SEMI[27] | 21.39/0.781 | 15.50/0.519 | 24.05/0.820 | 24.95/0.841 |
| RESCAN[25] | 23.09/0.830 | 24.86/0.783 | 27.46/0.864 | 27.14/0.869 |
| UMRL[26] | 23.92/0.883 | 24.85/0.835 | 27.73/0.929 | 29.59/0.922 |
| PreNet[28] | 24.03/0.872 | 25.75/0.861 | 31.64/0.949 | 30.86/0.926 |
| MSPFN[23] | 26.97/0.898 | 27.42/0.864 | 31.66/0.921 | 31.59/0.928 |
| **DPAFNet** | **28.58/0.916** | **28.54/0.889** | **32.84/0.944** | **31.89/0.932** |

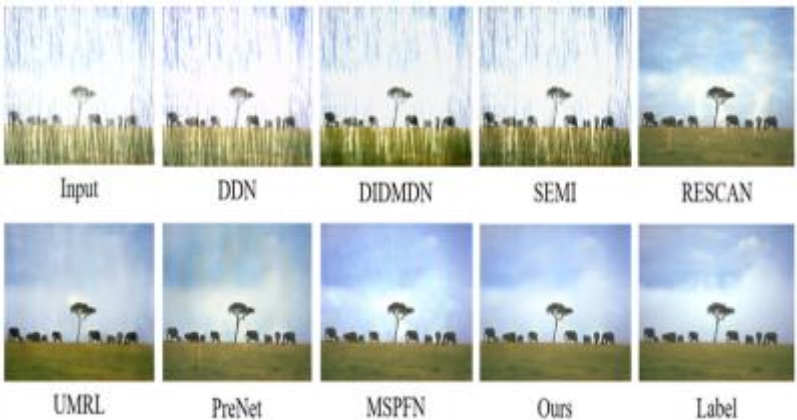

**Fig. 6** The comparison of the rain removal effect of the proposed method and multiple methods on the synthetic rain removal dataset Rain100H. Each synthetic rain image in Rain100 H contains rain streaks in multiple directions, which is difficult to recover.

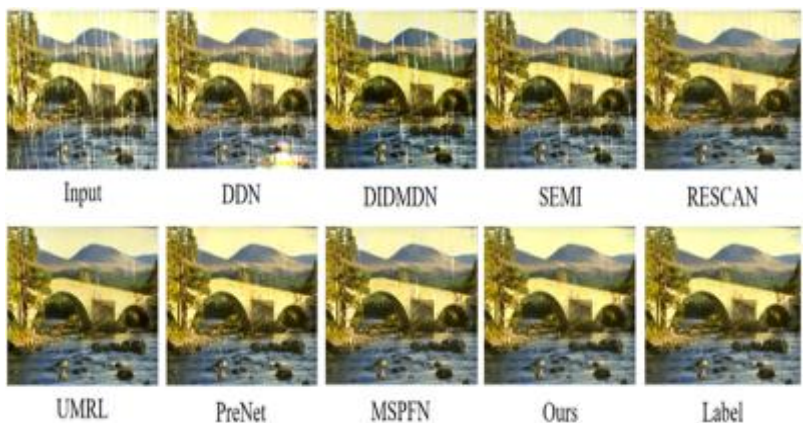

**Fig. 7** The comparison of the rain removal effect of the proposed method and multiple methods on the synthetic rain removal dataset Rain100L. Compared with other methods, our method has the best effect on removing rain streaks.

**Table 4** Comparisons on real-world dataset SPA-Data[61].

| Methods | PSNR | SSIM |
|---|---|---|
| Input | 34.15 | 0.927 |
| DSC[32] | 34.95 | 0.942 |
| GMM[33] | 34.30 | 0.943 |
| JCAS[34] | 34.95 | 0.945 |
| DerainNet[24] | 32.66 | 0.942 |
| DDC[28] | 34.70 | 0.934 |
| RESCAN[25] | 34.70 | 0.938 |
| JORDER_E[35] | 34.34 | 0.936 |
| DPAFNet | **34.67** | **0.951** |

### 4.3.2 Real Images

Due to the inevitable difference b between the synthetic rain image dataset and the images taken in real rainy days in nature, and in order to verify the practical application value of the model proposed in this paper for the single image rain task, we will propose the method on real rain images. The real rain image datasets we use are Internet-Data[27] and SPA-Data[31].At the same time, we compare the proposed method with a variety of methods, including DSC[32], GMM[33], JCAS[34], DerainNet[24], DDN[28], RESCAN[25], JORDER _ E[35]. We retrain the proposed model on the Rain800 dataset, and the comparison data used are from  . In addition, for SPA-Data, we provide quantitative analysis using PSNR and SSIM for comparison, as shown in Table 5; for Internet-Data, because it does not have a corresponding real benchmark clear image for quantitative analysis, we only provide a visual effect comparison diagram, as shown in Figure 8. Through Table 8 and Figure 8, it can be seen that our method has a good performance in removing rain on real rain images. It can remove rain streaks in real scenes and restore clear images.

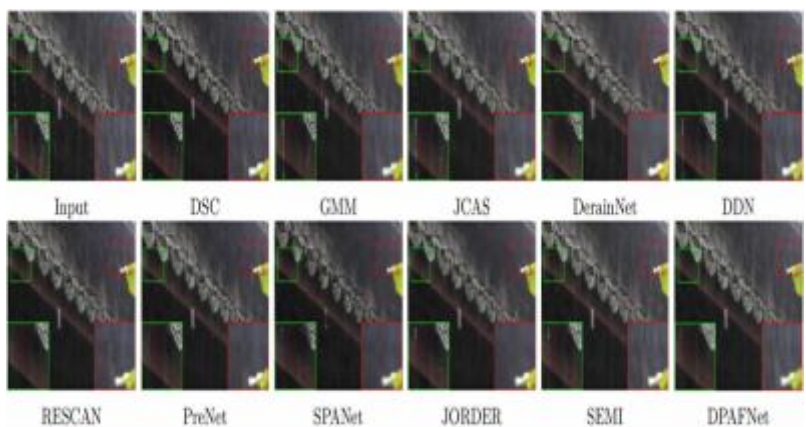

**Fig. 8** Deraining results on Intetnet-Data testing set. Best viewed when zoomed in and in color. It can be seen from the above image that our method can effectively remove the rain streaks in the natural rain image.

## 5 Conclusion

In this study, we propose a dual-branch attention fusion neural network for single image rain removal. Firstly, we propose a two-branch neural network. The two branches are the visual Transformer branch and the convolutional neural network branch, which can combine the long-distance dependence modeling ability and local feature modeling ability of the two branches. Then, we propose an attention-based feature fusion mechanism based on channel dimension, which can apply corresponding attention weights to the two branches to organically fuse the features of the two branches. Finally, the ablation experiment on the synthetic dataset proves the rationality of the proposed method, and the comparison experiment on the synthetic image and the real image proves the effectiveness of the proposed method.

**Author Contributions**: For Conceptualization, B.W.; methodology, B.W.; software, B.W.; validation, B.W., D.W. and Z.W.; formal analysis, L.Z.; investigation, B.W.; resources, B.W.; data curation, B.W.; writing—original draft preparation, B.W.; writing—review and editing, B.W.; visualization, B.W.; supervision, D.W.; project administration, D.W.; funding acquisition, L.Z. All authors have read and agreed to the published version of the manuscript.

**Funding:** This research was funded by Shandong Provincial Natural Science Foundation, China (Grant Number ZR2019BF022) and National Natural Science Foundation of China (Grant Number 62001272).

**Data Availability Statement:** All the datasets used for training the model of this paper are from Internet.

**Conflicts of Interest:** The authors declare no conflict of interest.